\documentclass[11pt,a4paper]{article}
\usepackage[hyperref]{acl2021}
\usepackage{times}
\usepackage{latexsym}

\usepackage{microtype}
\usepackage{graphicx}
\usepackage{pifont}
\newcommand{\cmark}{\ding{51}}%
\newcommand{\xmark}{\ding{55}}%
\usepackage{multirow, makecell}
\usepackage[utf8]{inputenc}
\DeclareFixedFont{\ttb}{T1}{txtt}{bx}{n}{8} 
\DeclareFixedFont{\ttm}{T1}{txtt}{m}{n}{8}  
\usepackage{color}
\definecolor{deepblue}{rgb}{0,0,0.5}
\definecolor{deepred}{rgb}{0.6,0,0}
\definecolor{deepgreen}{rgb}{0,0.5,0}
\usepackage{listings}
\newcommand\pythonstyle{\lstset{
language=Python,
basicstyle=\ttm,
morekeywords={self},              
keywordstyle=\ttb\color{deepblue},
emph={MyClass,__init__},          
emphstyle=\ttb\color{deepred},    
stringstyle=\color{deepred},
commentstyle=\ttb\color{deepgreen},
frame=tb,                         
showstringspaces=false
}}
\lstnewenvironment{python}[1][]
{
\pythonstyle
\lstset{#1}
}
{}
\newcommand\pythonexternal[2][]{{
\pythonstyle
\lstinputlisting[#1]{#2}}}
\newcommand\pythoninline[1]{{\pythonstyle\lstinline!#1!}}

\usepackage{colortbl}
\usepackage{balance}
\definecolor{mygray}{gray}{.9}
\usepackage[export]{adjustbox}

\newcommand\modelname{LEGOEval}
\newcommand\parlai{ParlAI}

\newcommand\blfootnote[1]{%
  \begingroup
  \renewcommand\thefootnote{}\footnote{#1}%
  \addtocounter{footnote}{-1}%
  \endgroup
}

\newcommand{\unimpara}[1]{\vspace{0.02in}\noindent{\textbf{#1}}}
\aclfinalcopy 

\title{LEGOEval: An Open-Source Toolkit for \\
Dialogue System Evaluation via Crowdsourcing}

\author{$^{\dagger}$Yu Li$^{*}$ \quad $^{\dagger}$Josh Arnold$^{*}$ \quad $^{\ddagger}$Feifan Yan \quad $^{\ddagger}$Weiyan Shi \quad $^{\ddagger}$Zhou Yu\\
~~\\
$^{\dagger}$Department of Computer Science, University of California, Davis\\
$^{\ddagger}$Department of Computer Science, Columbia University\\
  $^{\dagger}$\texttt{\{yooli, jarnold\}@ucdavis.edu} \\
  $^{\ddagger}$\texttt{\{fy2241, ws2634, zy2461\}@columbia.edu} \\}

\date{}

\begin{document}
\maketitle
\begin{abstract}
We present \modelname, an open-source toolkit that enables researchers to easily evaluate dialogue systems in a few lines of code using the online crowdsource platform, Amazon Mechanical Turk. Compared to existing toolkits, \modelname\ features a flexible task design by providing a Python API that maps to commonly used React.js interface components.
Researchers can personalize their evaluation procedures easily with our built-in pages as if playing with LEGO blocks. Thus, \modelname\ provides a fast, consistent method for reproducing human evaluation results. Besides the flexible task design, \modelname\ also offers an easy API to review collected data. 
\blfootnote{* Equal contribution.}
\blfootnote{Source code and documentation are available at \url{https://github.com/yooli23/LEGOEval}.}
\blfootnote{A demo video is available at \url{https://www.youtube.com/watch?v=Dg6mafRGOpg&ab_channel=JoshArnold}.}
\end{abstract}

\section{Introduction}
\label{section:intro}

\begin{table*}[htb!]
    \centering
    \begin{tabular}{lccccc}
    \hline
    \multirow{2}{*}{\textbf{Toolkit}}  & \multirow{2}{*}{\textbf{\modelname}}  & \multirow{2}{*}{\shortstack[c]{\textbf{DialCrowd}\\ \cite{lee2018dialcrowd}}} &  \multirow{2}{*}{\shortstack[c]{\textbf{\parlai}\\ \cite{miller2017parlai}}} & \multirow{2}{*}{\textbf{Mephisto}} \\ \\
    \hline
    Sample Templates    & \cmark &  \cmark & \cmark & \cmark \\
    Flexible Interface Design  & \cmark & \xmark & \xmark &  \xmark \\
    Branching Logic & \cmark & \xmark & \xmark &  \xmark \\
    Plug \& Play     &  \cmark & \xmark &  \xmark &  \cmark \\
    Data Reviewing Tool & \cmark & \cmark  & \xmark & \cmark \\
    \hline
    \end{tabular}
    \caption{Comparison of related crowdsourcing tools. ``Sample Templates'' indicates that the tool has evaluation examples that are commonly used. ``Flexible Interface Design'' indicates that the evaluation interface can be fast and easily modified. ``Branching Logic'' means the tool supports different interfaces in a same task.``Plug \& Play'' means that the tool can be used out of the box.}
    \label{tab:comparison}
\end{table*}

As dialogue systems are becoming an increasingly trending topic, the need for standardized and reliable evaluation procedures has grown significantly. Typically, the evaluation of dialogue systems is accomplished by the use of both automatic metrics \cite{papineni2002bleu, lin2004rouge, lavie2007meteor} and human evaluation \cite{serban2016building, park2018hierarchical}. Automatic metrics are reliable measurements, but common automatic metrics correlate weakly with human judgment \cite{liu2016not, lowe2017towards, gu2020perception}. Thus, human evaluation has become a primary method for dialogue system evaluation. Previously, researchers invited participants to the lab to physically interact with dialogue systems; recently, the popular approach is crowdsourcing using platforms such as Amazon Mechanical Turk (AMT) \cite{deriu2021survey, eskenazi2013crowdsourcing}.


However, human evaluation via crowdsourcing presents its own challenges, being both expensive and time-intensive. Specifically, human evaluation requires a huge engineering effort to develop the interface and deploy the task on crowdsourcing platforms. The front-end interfaces can be difficult to set up: the crowdworkers need to be properly instructed, and the tasks need to be prepared to reflect real-world environment as closely as possible. Furthermore, one needs to take into account the high variability of user behaviour especially in crowdsourced environments \cite{deriu2021survey}. It was shown that even different phrasings can result in weaker levels of agreement \cite{li2019acute}. Thus, it is not trivial to reproduce the human evaluation results from scratch.

To address these problems, we present \modelname, an open-source toolkit that enables researchers to easily build and deploy their human evaluation tasks on AMT in one click. \modelname\ supports representative human evaluation tasks, such as \textit{static evaluation}, where crowdworkers are asked to rate sampled dialogues, and \textit{interactive evaluation}, where crowdworkers interact with two systems and evaluate their responses \cite{finch2020towards, adiwardana2020towards}. Furthermore, researchers are also able to customize their own human evaluation procedures easily with \modelname.

Existing tools typically provide rigid human evaluation templates. For example, DialCrowd \cite{lee2018dialcrowd} follows the speech synthesis evaluation toolkit \cite{parlikar2012testvox} and provides a small number of standard evaluation experiments, however, researchers have to manually create the web services and then post the evaluation task on AMT. \citet{sedoc2019chateval} developed ChatEval, which posts a response comparison task \cite{otani2016irt} on AMT. It is only effective for specific dialogue systems and is not generalizable. The widely used toolkit \parlai\ \cite{miller2017parlai} supports  crowdsourcing tasks on AMT for the models built by \parlai. However, \parlai\ also requires additional engineering efforts to incorporate an external model or modify the evaluation interface. To our best knowledge, the most similar tool to ours is Mephisto \footnote{https://github.com/facebookresearch/Mephisto}, a crowdsourcing tool in an early alpha release expanded from \parlai\. From our experience, however, Mephisto has a steeper learning curve and is currently not suited for easily customizing and launching simple tasks.

Compared to these existing tools, our toolkit features a flexible interface design with a plug and play fashion, as shown in Table \ref{tab:comparison}. Researchers can build their personalized human evaluation task flow with our library of Python classes, including a chatbot interface, an instruction page, and various survey formats. The task building process is similar to playing a LEGO game. Furthermore, \modelname\ makes it easy to share tasks with others, thereby making it easy to reproduce human evaluation results. Additionally, \modelname\ provides a straight-forward way to persist, retrieve, and review collected data, thus helping researchers process their results more efficiently.

In this paper, we present \modelname\ in the following order: first, we describe the design and architecture in Section \ref{section:architecture},  then we provide code snippets showing how to build the personalized task page in Section \ref{section:usage}. Finally in section \ref{section:exp}, we reproduce past experiments using \modelname.

\begin{figure*}
    \centering
    \includegraphics[trim={0 0 0 0},clip,width=16cm]{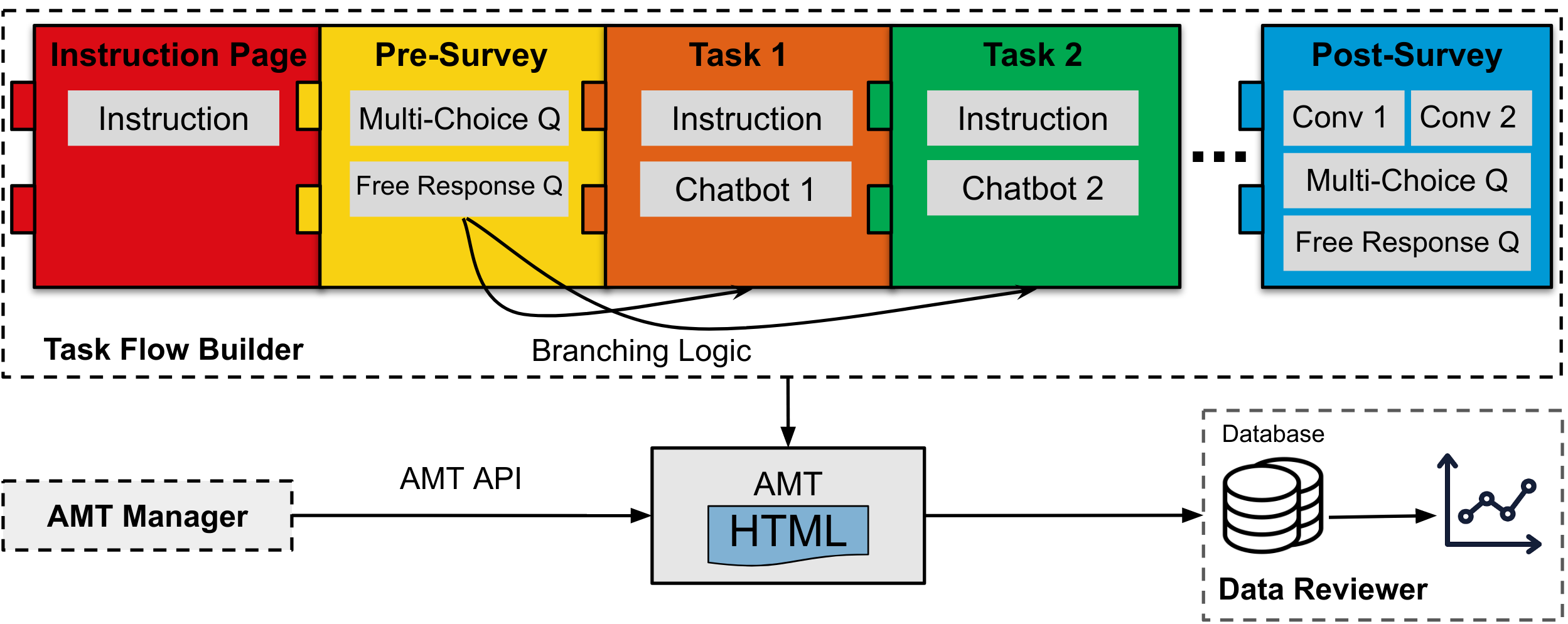}
    \caption{Design of \modelname, the task building process is similar to playing a LEGO game, the pages can be viewed as LEGO blocks in the task flow builder. The developer can also add \textit{branching logic} in their task with the lambda function to show different task flows to different crowdworkers at runtime.}
    \label{fig:archi}
    \vspace{-3mm}
\end{figure*}

\section{\modelname\ Toolkit}
\label{section:architecture}
\modelname\ is an open-source Python-based toolkit.
As shown in Figure \ref{fig:archi}, \modelname\ includes three modules: (1) a task flow builder (Section \ref{subsec:tfb}) for designing the human evaluation task, (2) an AMT manager (Section \ref{subsec:as}) that automatically deploys the evaluation task on AMT, and (3) a data reviewer API  (Section \ref{subsec:dp}) that retrieves and formats data collected on AMT. We describe each module in the following sections.

\subsection{Task Flow Builder}
\label{subsec:tfb}
The task flow builder generates the interface and flow for different human evaluation tasks  by compiling a list of \textit{pages}. The pages can be viewed as LEGO blocks: we can snap pages together to easily customize the evaluation task flow. Furthermore, our toolkit and LEGO-style design makes it easy to share tasks: instead of sharing an entire web application, researchers can simply share their tasks with a few lines of code in-order to reproduce their evaluation procedure.

We have also provided common human evaluation procedures in \modelname\, for instance, \textit{static evaluation}, where crowdworkers are asked to rate sampled dialogues and \textit{interactive evaluation}, where a crowdworker interacts with multiple dialogue systems and evaluates the responses. Researchers can easily integrate their models, and customize their task flow using \modelname.

\unimpara{Page} As the name suggests, a page in \modelname\ is a single web page with a specific functionality (e.g. displaying instructions, presenting a survey). Pages are designed to be independent of one-another, preventing any complex dependencies from occurring. Furthermore, each page is defined by a single React.js file and mapped to a simple Python wrapper class. We have provided a pool of pages that are commonly used in human evaluation tasks, including an instruction page (to display task instructions), an interactive chatbot page, and various survey pages. Beyond the built-in pages and their parameters, researchers also have the flexibility to customize a page or its logic by simply editing a single React.js file.

\unimpara{Page Customization} Customizing the front-end display of a page is as simple as editing a single React.js file. Researchers can easily add an image, or re-arrange the order of the user-interface elements. With React.js, one can also edit a page's CSS, achieving complex front-end layouts if needed. Lastly, if researchers need to modify the logic/functionality of a component, they can also do it from the same React.js file through the use of \texttt{state}. We further describe how \texttt{state} works in \modelname\ in the next paragraph.

\unimpara{State} In \modelname, researchers are able to easily customize the functionality of a page with the idea of \texttt{state}. \texttt{State} solves the dilemma that often occurs when designing a human evaluation task: specifically, how one should design the data flow between the back-end and the front-end. To address this issue of data-flow design, we implement the idea of a shared-state between the front-end and the back-end. When the back-end modifies the state of a task, the state is automatically updated on the front-end, and vice versa. Furthermore, any changes made to the state are automatically persisted in a Postgres-SQL database, making data persistence and retrieval incredibly simple. Thus, when modifying the functionality of React.js page, the researcher only needs to set key-value pairs in the front-end React.js state-dictionary. Their key-value pairs will automatically be persisted in a database and synced to the backend. Although not always necessary to modify, \modelname\ also features a main loop function on the backend, found in \texttt{main\_loop.py}, that is called each time the \texttt{state} is modified from the front-end. In the main loop, the researcher can respond to any front-end changes from the backend. For example, when a crowdworker sends a chat message on the front-end, the main loop provides the backend an opportunity to provide a response. We provide detailed documentation on the main loop in our GitHub repository. Thus, with \texttt{state}, the researcher can easily save or pass data between the backend and their front-end React.js file, allowing them to flexibly implement any needed functionality.

\unimpara{Building a Task Flow} To build an evaluation task in \modelname, we just need to assemble the pages in a similar fashion to building LEGO. An evaluation task usually consists of a multi-phase flow, e.g., displaying task instructions, then a survey, etc. Each phase corresponds to an individual page. The developer can add the desired pages in a sequence, and \modelname\ will automatically display each page in order one at a time. Additionally, \modelname\ supports branching logic to show different task flows to different crowdworkers, determined at runtime. We will describe branching logic in more detail in Section~\ref{sec: additional functions}.

\subsection{AMT Manager}
\label{subsec:as}
Once the task flow is created, the task flow builder will automatically generate the necessary files that can be embedded in an AMT task. Our AMT manager follows the Mechanical Turk manager pipeline in \parlai\ \cite{miller2017parlai} and launches the evaluation task on AMT by embedding the generated interface using an iFrame. Researchers can edit the AMT configuration file which contains the AMT task settings, including the reward for each Human Intelligence Task (HIT), the number of HITs to launch, the task title, etc. When researchers launch the task, the AMT manager will automatically build and deploy a web application on the cloud application platform, Heroku \footnote{https://www.heroku.com/}, and then post the evaluation task on AMT using the AMT API. Our AMT manager inherits necessary functions from the Mechanical Turk manager in \parlai , for example, checking crowdworker's qualification and optionally limiting the number of hits for each crowdworker. The AMT manager also supports automatic data validation. For instance, when crowdworkers finish a HIT, the manager can check the quality of the collected data via metrics defined by the researcher (e.g., the dialogue length or the rating variance in the rating question component), and then it will approve or deny the reward and bonus, depending on if the results pass the data validation test.

\subsection{Data Reviewer}
\label{subsec:dp}
\modelname\ also simplifies the process to review the collected data. Different from \parlai\ \cite{miller2017parlai} which saves raw data locally, we create an individual Heroku Postgres add-on as the database of the web application when researchers launch a new task. We will create separate databases when there are multiple evaluation tasks running on AMT at the same time. Collected data will be saved in the corresponding add-on database. The data can be read locally using our Python API. Storing the data on the cloud prevents potential accidents, such as locally deleting the data. Since there are various types of data that can be collected in \modelname\ (e.g., conversations, ratings, and free response questions), we organize the data according to the data type so that they can be easily reviewed and processed for further research using a Python API we wrote.

\subsection{Additional Functions}
\label{sec: additional functions}
As mentioned earlier, \modelname\ is a flexible toolkit to easily implement human evaluation tasks. Specifically, it allows branching logic to display pages dynamically at run-time, as well as guaranteed data collection of a fixed size, which are two important features in dialogue human evaluation.

\unimpara{Branching Logic} \modelname\ is unique in the way that our pages are instantiated at run-time. To implement branching logic, we offer a special lambda function that is called at runtime. Each lambda function takes as an argument the current \texttt{state} of a task. Researchers can build \textit{branching and conditional logic} in their tasks by using lambda functions. For example, a researcher can easily define logic to skip a certain page based upon a crowdworker's previous answer to a survey. We show an example of branching logic implementation in Section \ref{section:usage}.

\unimpara{Assigning Tasks to Workers} Another common problem faced with crowd-sourced tasks is collecting data in a distributed, sampled fashion. For example, if you have 100 conversations, you might want each conversation to be rated \textbf{exactly} $X$ times, each time by a different crowd-worker. Because crowd-workers can start a task and then give up half-way through, building the logic to assign HITS in the aforementioned way can be time-consuming. To achieve this in \modelname\, we provide a Python wrapper class, named \texttt{Data Assigner}. When using the Data Assigner class, you simply pass in a list of json seralizable data (dictionaries, arrays, strings, etc) and specify how many times you want to collect each data point. After that, \modelname\ will automatically distribute the data to be randomly assigned to different workers until each data point has been successfully collected $X$ times.

\begin{figure*}
    \centering
    \includegraphics[trim={0 0 0 0},clip,width=16cm]{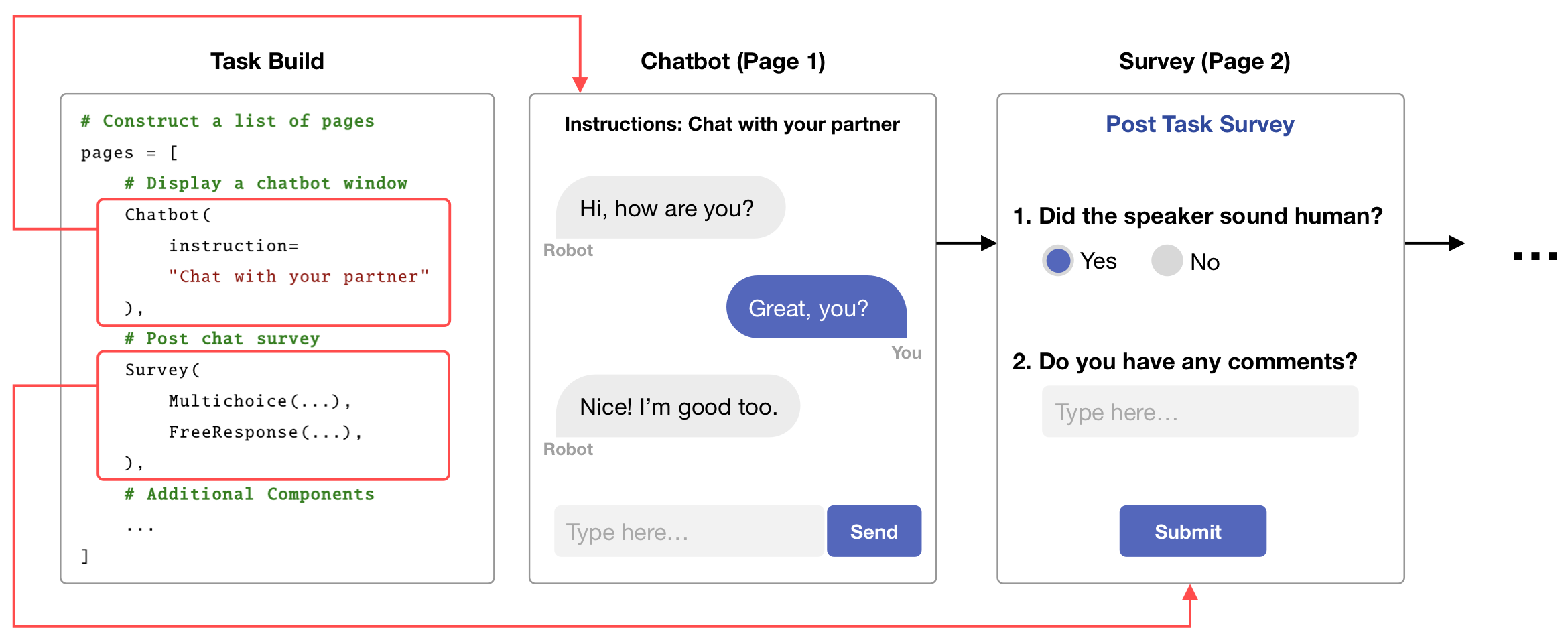}
    \caption{An example how the task build maps to a generated evaluation task in  Section \ref{section:usage}.}
    \label{fig:usageExample}
    \vspace{-3mm}
\end{figure*}

\section{Toolkit Usage}
\label{section:usage}

In this section, we provide a simple example of what the researcher needs to do to create a task.

\unimpara{Typical Usage} Researchers would first design their high-level task flow on paper, (e.g, instruction page, then a pre survey, ..., and finally, a post survey). After this, researchers can add their different pages to their task flow by editing the \texttt{build.py} file and initializing a list of our provided Python classes. Some pages, such as the survey page, have a high level of customization, where developers can specify what types of questions they want to display on the survey, and the questions' relative order. Next, researchers can test their task locally by running \texttt{server.py} and navigating to their localhost. If they are happy with the results, researchers can launch their task on MTurk with one command: \texttt{launch\_hits.py}. If researchers want to make a few tweaks, such as using a custom font for the instructions, they can easily edit a single React.js file and override any necessary CSS in typically one line of code. A strong benefit of our platform is that it is very fast for researchers to plug and play typical evaluation tasks. For further detail, see Figure~\ref{fig:usageExample} for an example.

\unimpara{Advanced Usage} As previously mentioned, it is possible to add branching logic via the use of lambda functions. Firstly, the researcher must define a function that takes as an input the state dictionary. The state dictionary contains information representing the current state of the task and any persisted data. Thus, based upon the data collected so far, the researcher can decide which page to instantiate and return. The researcher adds the \textit{LambdaFunction} object passing in their lambda function. During run time, when the \textit{LambdaFunction} object is popped from the task flow list, our framework will call the function by passing in the current state dictionary and return the determined page to display.
 
\pythonexternal{code_example3.py}

\section{Experiments}
\label{section:exp}
To demonstrate the effectiveness of \modelname\ in setting up dialogue system human evaluation tasks, we reproduce a set of crowdsourced experiments from the BlenderBot paper \cite{roller2020recipes}, a state-of-the-art open-domain chatbot. Crowdsourced experiments in BlenderBot include two steps: (1) collecting human-bot conversations via crowdworkers, (2) ACUTE-Eval \cite{li2019acute} between two models, where crowdworkers are asked to make pairwise evaluations of complete dialogues. We implement both crowdsourced experiments in approximately 20 lines of Python code with  \modelname, indicating that it is easy to implement different types of human evaluation tasks with our toolkit in a  plug and play fashion.

\subsection{Human-Bot Data Collection}
\begin{figure*}
    \centering
    \includegraphics[trim={0 0 0 0},clip,width=16cm]{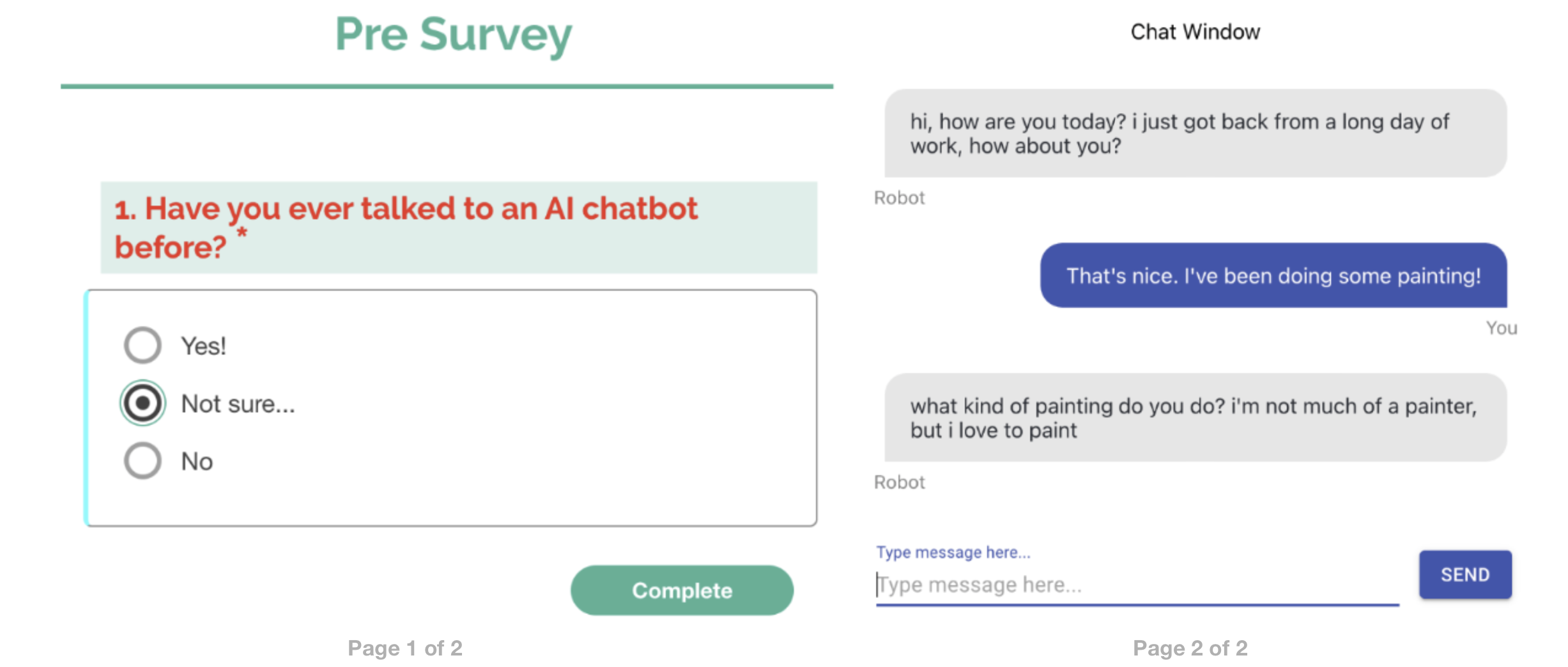}
\caption{\label{fig:exp1_interface} The interface of the human-bot conversation task with the Blender Model. In the task, the survey is shown first, and then the chat window. However, we show the pages side by side for convenience.}
    \vspace{-3mm}
\end{figure*}

Following \citet{roller2020recipes} and \citet{adiwardana2020towards}, we build a task to collect human-bot conversations on AMT for the 90M BlenderBot model with \modelname. We simply assemble a pre-survey question component in the first page and a chatbot component in the second page. The generated interface is shown in Figure \ref{fig:exp1_interface}. We build the whole task from scratch and post it on AMT in several minutes. Then we collect 20 conversations following the settings in \citet{roller2020recipes}. One example from our collected conversations is shown in \autoref{appendix:example_conversation}. It shows that our toolkit can collect human-bot conversations properly.
 
\subsection{BlenderBot (2.7B) vs. Meena}
To demonstrate \modelname's capability to support different dialogue system human evaluation tasks, we also reproduce ACUTE-Eval \cite{li2019acute} between BlenderBot \cite{roller2020recipes} and Meena \cite{adiwardana2020towards}. ACUTE-Eval requires human annotators to compare multi-turn conversations between different dialogue systems. Following \citet{roller2020recipes} and \citet{li2019acute}, we consider two evaluation questions:

\begin{itemize}
    \item Engagingness question: ``Who would you prefer to talk to for a long conversation?''
    \item Humanness question: ``Which speaker sounds more human?''
\end{itemize}

\begin{figure}[htb!]
\vspace{-2mm}
\centering
{
\includegraphics[trim={0 0cm 0 0cm},clip,width=7.7cm]{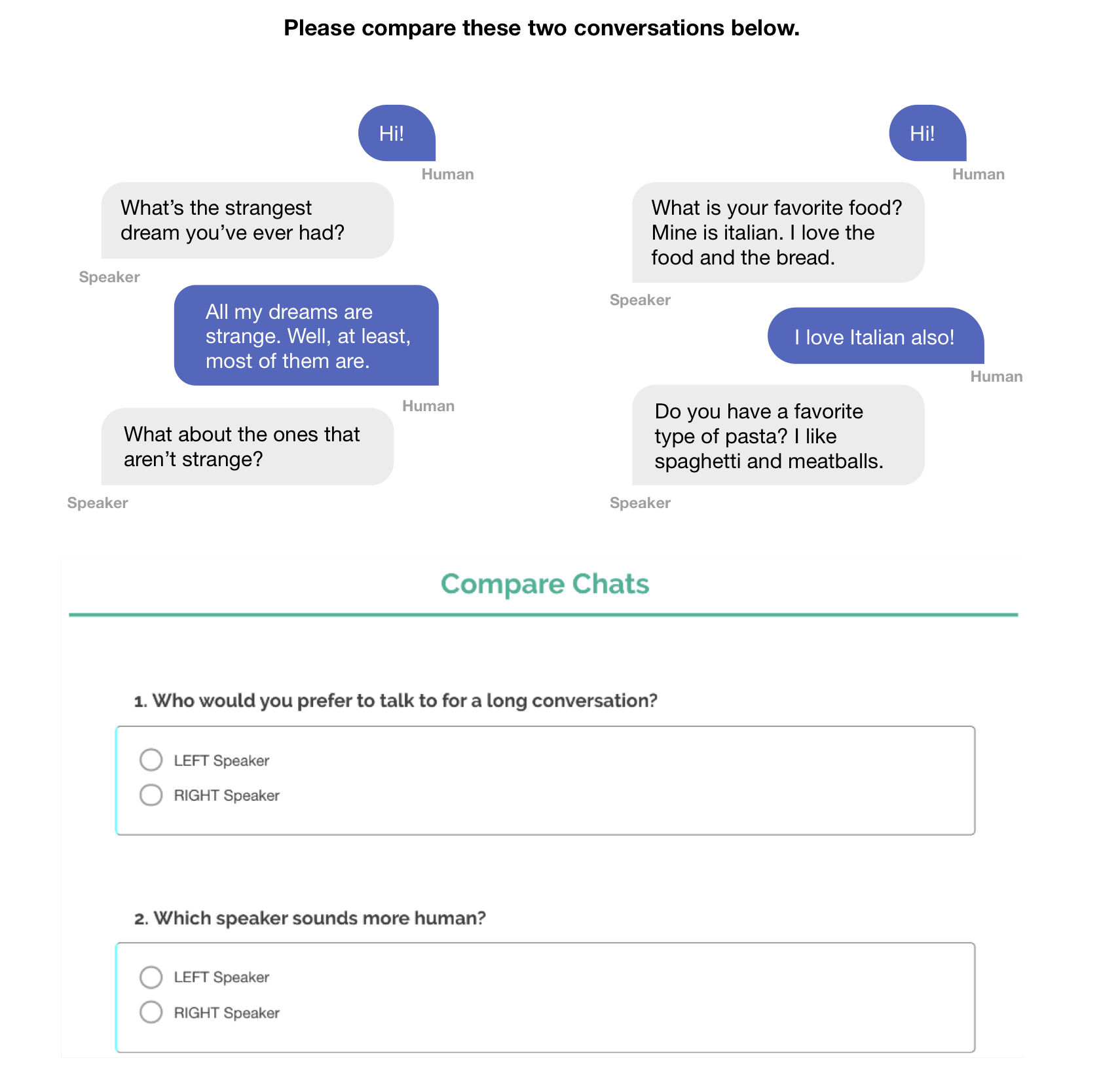}
}
\caption{\label{fig:exp2_interface} Interface of ACUTE-Eval between the BlenderBot and Meena. The conversations have been truncated to save display-space.}
\vspace{-5mm}
\end{figure}

As shown in Figure \ref{fig:exp2_interface}, we use a chat history comparison page that features a single choice survey question. Crowdworkers are instructed to compare two dialogues and answer two questions below. We collect 114 data points, the results is shown in Table \ref{tab:exp2_results}. The results (0.72 vs. 0.28 for engagingness, 0.68 vs. 0.32 for humanness) are close to the results in \citet{roller2020recipes} (0.75 vs. 0.25 for engagingness, 0.65 vs. 0.35 for humanness). It demonstrates that with our toolkit, we can reproduce the human evaluation results in other works quickly.

\begin{table}[htb!]
\centering
\begin{tabular}{l|cc}
    \hline
    \textbf{Model}& \textbf{BlenderBot} & \textbf{Meena}\\
    \hline
    Engagingness & 0.72 (0.75) & 0.28 (0.25)\\
    \hline
    Humanness & 0.68 (0.65) & 0.32 (0.35)\\
    \hline
\end{tabular}
\caption{Human-Chat ACUTE-Eval of engagingness and humanness between BlenderBot (2.7B) and Meena, numbers represent the percentage of people choose the model. Numbers in brackets are the results in \citet{roller2020recipes}}
\label{tab:exp2_results}
\end{table}

\section{Conclusion and Future Work}
We introduce \modelname, an open-source Python-based toolkit that allows researchers to easily develop human evaluation tasks for dialogue systems on AMT in a LEGO plug-and-play fashion. \modelname\ provides a variety of commonly-used React.js components as building blocks for researchers to use. We have shown that it is straightforward to customize different types of human evaluation procedures for dialogue systems in a few lines of code, create new components by leveraging the \textit{shared state} between the front-end and back-end, and reproduce human evaluation results in other works effortlessly. With \modelname, we hope to provide a simple and flexible way to evaluate dialogue systems. For future work, we plan to add more human evaluation procedure templates and React.js components. We further want to extend the tool to support human-human interaction on AMT, making it work not only for evaluation, but also for dialogue data collection.
\bibliographystyle{acl_natbib}
\bibliography{anthology,custom}

\clearpage
\appendix

\section{Example conversation}
\label{appendix:example_conversation}
Table \ref{tb:dialog example} shows an example conversation we collected in the human-bot data collection experiment.

\begin{table}[htb!]
\small
\centering
\centering
\begin{tabular}{lp{58mm}}
\hline
\rowcolor{white} \textbf{Role} & \bf Utterance\\
\hline
\rowcolor{mygray} Human   &  Hi  \\
\rowcolor{white} Bot   & Hi, how are you today? I just got back from a long day of work, how about you?\\
\rowcolor{mygray} Human   &  I have had a day off today. Nothing much to do.  \\
\rowcolor{white} Bot   & Oh, I'm sorry to hear that what do you like to do in your spare time?\\
\rowcolor{mygray} Human   & I like to read a lot. I also like to go to the movies.\\
\rowcolor{white} Bot   & I love to read as well. What kind of books do you read? I've never been to a movie\\
\rowcolor{mygray} Human   & I like the Harry Potter series and the Lord of the Rings trilogy very much. \\
\rowcolor{white} Bot   & I haven't read any of those books, but I have heard good things about them, do you have any other hobbies? \\
\rowcolor{mygray} Human   & I like to go walking in the countryside. Do you like nature?\\
\rowcolor{white} Bot   & I do like nature, but not as much as I would like to spend time in the wild\\
\rowcolor{mygray} Human   & Do you enjoy camping? I really love that.\\
\rowcolor{white} Bot   & No, I don't think I could do that. I think I would get sick of it.\\
\rowcolor{mygray} Human   &  But it's so nice to be outside interacting with nature.\\
\rowcolor{white} Bot   & That's true, maybe I'll try it one day, it sounds like a lot of fun.\\
\rowcolor{mygray} Human   &  You can cook outside too. Do you like cooking?\\
\rowcolor{white} Bot   & I do love to cook, I do it all the time. What is your favorite thing to cook?\\
\rowcolor{mygray} Human   &  I like to bake a lot. I make really good chocolate chip cookies.\\
\hline
\end{tabular}
\caption{\label{tb:dialog example} One example conversation between BlenderBot and human collected by \modelname.}
\end{table}

\label{rule-partitioning-explan}

\end{document}